\documentclass[10pt,twocolumn,letterpaper]{article}

\usepackage{cvpr}
\usepackage{times}
\usepackage{epsfig}
\usepackage{graphicx}
\usepackage{amsmath}
\usepackage{amssymb}
\usepackage{subfigure}

\usepackage[breaklinks=true,bookmarks=false]{hyperref}

\cvprfinalcopy 

\newcommand*{\affaddr}[1]{#1} 
\newcommand*{\affmark}[1][*]{\textsuperscript{#1}}
\newcommand*{\email}[1]{ \small \texttt{#1}}
\renewcommand{\thefootnote}{\fnsymbol{footnote}}

\def\cvprPaperID{****} 

\begin{document}

\title{An End-to-End Network for Panoptic Segmentation}
\setcounter{footnote}{-1}
\author{%
Huanyu Liu\affmark[1]\footnotemark[2], Chao Peng\affmark[2], Changqian Yu\affmark[3]\footnotemark[2], Jingbo Wang\affmark[4]\footnotemark[2], Xu Liu\affmark[5]\footnotemark[2], Gang Yu\affmark[2], Wei Jiang\affmark[1]\\
\affaddr{\affmark[1]Zhejiang University},  \affaddr{\affmark[2]Megvii Inc. (Face++)},
\affaddr{\affmark[3]Huazhong University of Science and Technology} \\
\affaddr{\affmark[4]Peking University},
\affaddr{\affmark[5]The University of Tokyo}\\
\email{\{liuhy, jiangwei\_zju\}@zju.edu.cn},
\email{mikejay0520@163.com},
\email{changqian\_yu@hust.edu.cn}\\
\email{wangjingbo1219@pku.edu.cn},
\email{liuxu@kmj.iis.u-tokyo.ac.jp},
\email{yugang@megvii.com}\\
}

\maketitle

\renewcommand{\thefootnote}{\fnsymbol{footnote}}
\footnotetext[2]{This work was done during an internship at Megvii Inc.}

\begin{abstract}

Panoptic segmentation, which needs to assign a category label to each pixel and segment each object instance simultaneously, is a challenging topic. Traditionally, the existing approaches utilize two independent models without sharing features, which makes the pipeline inefficient to implement. In addition, a heuristic method is usually employed to merge the results. However, the overlapping relationship between object instances is difficult to determine without sufficient context information during the merging process. To address the problems, we propose a novel end-to-end Occlusion Aware Network (OANet) for panoptic segmentation, which can efficiently and effectively predict both the instance and stuff segmentation in a single network.
Moreover, we introduce a novel spatial ranking module to deal with the occlusion problem between the predicted instances. Extensive experiments have been done to validate the performance of our proposed method and promising results have been achieved on the COCO Panoptic benchmark.

\end{abstract}

\section{Introduction}
Panoptic segmentation~\cite{kirillov2018panoptic} is a new challenging topic for scene understanding. The goal is to assign each pixel with a category label and segment each object instance in the image. In this task, the stuff segmentation is employed to predict the  amorphous regions (noted \emph{Stuff}) while the instance segmentation~\cite{he2017mask} solves the countable objects (noted \emph{Thing}). Therefore, this task can provide more comprehensive scene information and can be widely used in autonomous driving and scene parsing.

\begin{figure}
	\subfigure{
		\begin{minipage}{1.0\textwidth}
		\label{main figure}
		\includegraphics[width=0.5\linewidth]{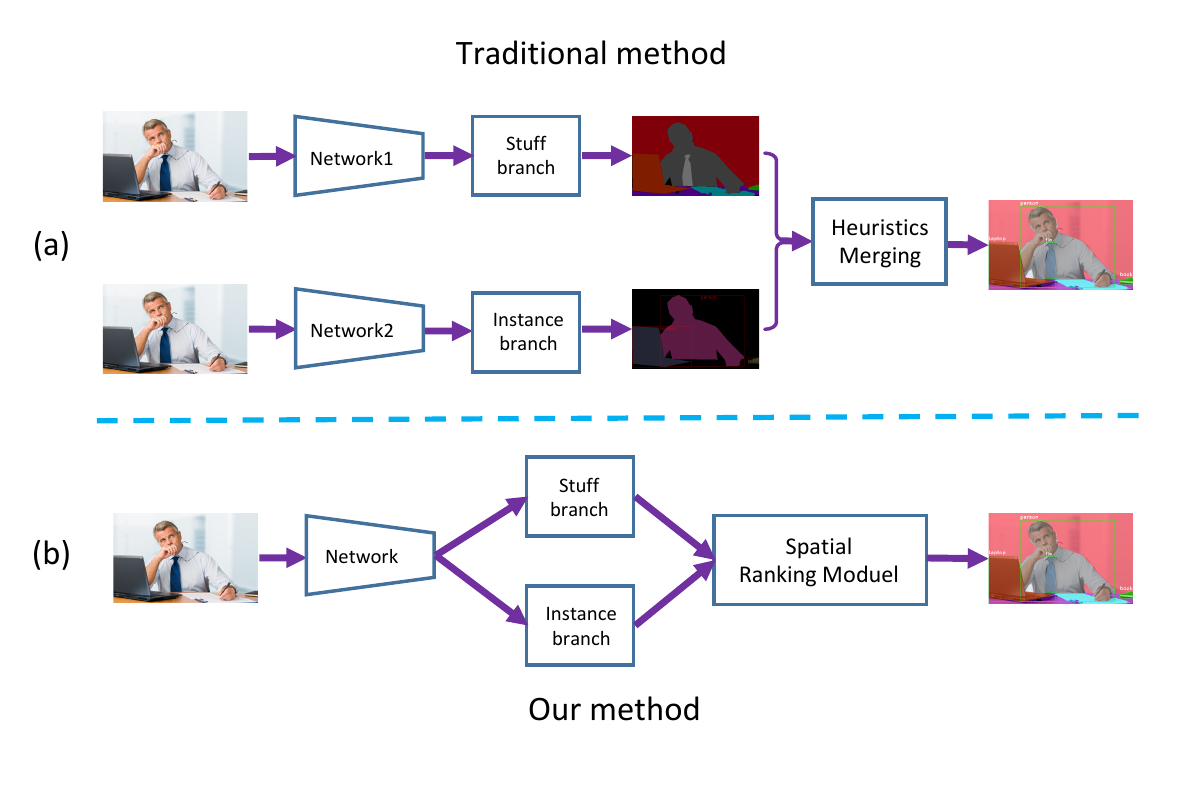}
		\end{minipage}
	}
	\caption{An illustration of our end-to-end network in contrast to the traditional method. Traditional methods~\cite{kirillov2018panoptic} train two subnetwork and do heuristics merge. Our method can train a single network for two subtasks, and realize a learnable fusion approach.}
	\label{fig:introduction_contrast}
\end{figure}

Previous panoptic segmentation algorithms~\cite{kirillov2018panoptic} usually contain three separated components: the instance segmentation block, the stuff segmentation block and the merging block, as shown in Figure~\ref{fig:introduction_contrast} (a).  Usually in these algorithms, the instance and stuff segmentation blocks are independent without any feature sharing. This results in apparent computational overhead. Furthermore, because of the separated models, these algorithms have to merge the corresponding separated predictions with post-processing. However, without the context information between the stuff and thing, the merge process will face the challenge of overlapping relationships between instances and stuff. Exactly as mentioned above, with the three separate parts, it is hard to apply this complex pipeline to the industrial application.

In this paper, we propose a novel end-to-end algorithm shown in Figure~\ref{fig:introduction_contrast} (b). As far as we know, this is the first algorithm which can deal with the issues above in an end-to-end pipeline. More specifically, we incorporate the instance segmentation and stuff segmentation into one network, which shares the backbone features but applies different head branches for the two tasks. During the training phase, the backbone features will be optimized by the accumulated losses from both the stuff and thing supervision while the head branches will only fine-tune on the specific task.

To solve the problem of overlapping relationship between object instances, we also introduce a new algorithm called \emph{Spatial Ranking Module}. This module learns the ranking score and offers an ordering accordance for instances.

In general, we summarize the contributions of our algorithm as follows:
\begin{itemize}
\item We are the first to propose an end-to-end occlusion aware pipeline for the problem of panoptic segmentation.
\item We introduce a novel spatial ranking module to address the ambiguities of the overlapping relationship, which commonly exists in the problem of panoptic segmentation.
\item We obtain state-of-the-art performance on the COCO panoptic segmentation dataset.
\end{itemize}

\section{Related Work}

\subsection{Instance Segmentation}

There are currently two main frameworks for instance segmentation, including the proposal-based methods and segmentation-based methods. The proposal-based approaches~\cite{dai2016instance, he2017mask, li2017fully, li2018detnet, liu2018path, liu2016multi, peng2018megdet} first generate the object detection bounding boxes and then perform mask prediction on each box for instance segmentation. These methods are closely related to object detection algorithms such as Fast/Faster R-CNN and SPPNet~\cite{girshick2014rich, he2014spatial, ren2015faster}.
Under this framework, the overlapping problem raises due to the independence prediction of distinct instances. That is, pixels may be allocated to wrong categories when covered by multiple masks.
The segmentation-based methods use the semantic segmentation network to predict the pixel class, and obtain each instance mask by decoding the object boundary~\cite{kirillov2017instancecut} or the custom field~\cite{bai2017deep, de2017semantic, liu2017sgn}. Finally, they use the bottom-up grouping mechanism to generate the object instances. RNN method was leveraged to predict a mask for each instance at a time in~\cite{ren2017end, romera2016recurrent, zhang2015monocular} .

\subsection{Semantic Segmentation}

Semantic segmentation has been extensively studied, and many new methods have been proposed in recent years. Driven by powerful deep neural networks~\cite{he2016deep, krizhevsky2012imagenet, simonyan2014very, szegedy2015going}, FCN~\cite{long2015fully} successfully applied deep learning to pixel-by-pixel image semantic segmentation by replacing the fully connected layer of the image classification network with the convolutional layer.

The encoder-decoder structure such as UNet~\cite{badrinarayanan2015segnet, ronneberger2015u, yu2018bisenet, yu2018learning, zhang2018exfuse} can gradually recover resolution and capture more object details. Global Convolutional Network~\cite{peng2017large} proposes large kernel method to relieve the contradiction between classification and localization. DFN~\cite{yu2018learning} designs a channel attention block to select feature maps.
DeepLab~\cite{chen2018deeplab, chen2018encoder} and PSPNet~\cite{zhao2017pyramid} use atrous spatial pyramid pooling or spatial pyramid pooling to get multi-scale context.
Method of~\cite{yu2015multi} uses dilated convolution to enlarge field of view. Multi-scale features were using to obtain sufficient receptive field in~\cite{chen2016attention, hariharan2015hypercolumns, xia2016zoom} .

Related datasets are also constantly being enriched and expanded. Currently, there are public datasets such as VOC~\cite{everingham2015pascal}, Cityscapes~\cite{cordts2016cityscapes}, ADE20K~\cite{zhou2017scene}, Mapillary Vistas~\cite{neuhold2017mapillary}, and COCO Stuff~\cite{caesar2018coco}.

\subsection{Panoptic Segmentation}
Panoptic Segmentation task was first proposed in~\cite{kirillov2018panoptic} and the research work for this task is not too much. A weakly supervised model that jointly performs semantic and instance segmentation was proposed by~\cite{li2018weakly}. It uses the weak bounding box annotations for ``thing" classes, and image level tags for ``stuff" classes. JSIS-Net~\cite{de2018panoptic} proposes a single network with the instance segmentation head~\cite{he2017mask} and the pyramid stuff segmentation head~\cite{zhao2017pyramid}, following heuristics to merge two kinds of outputs. Li et al.~\cite{li2018attention} propose AUNet that can leverage proposal and mask level attention and get better background results.

\subsection{Multi-task learning}
Panoptic segmentation can also be treated as multi-task learning problem. Two different task can be trained together through strategies~\cite{kendall2018multi, misra2016cross}. UberNet~\cite{kokkinos2017ubernet} jointly handles low-,
mid-, and high-level vision tasks in a single network, including boundary detection, semantic segmentation and normal estimation. Zamir et al.~\cite{zamir2018taskonomy} build a directed graph named taskonomy, which can effectively measure and leverage the correlation between different visual tasks. It can avoid repetitive learning and enable learning with less data.
\begin{figure*}
	\centering
	\subfigure{
		\begin{minipage}{1.0\textwidth}
		\label{main figure}
		\centering
		\includegraphics[width=1.0\linewidth]{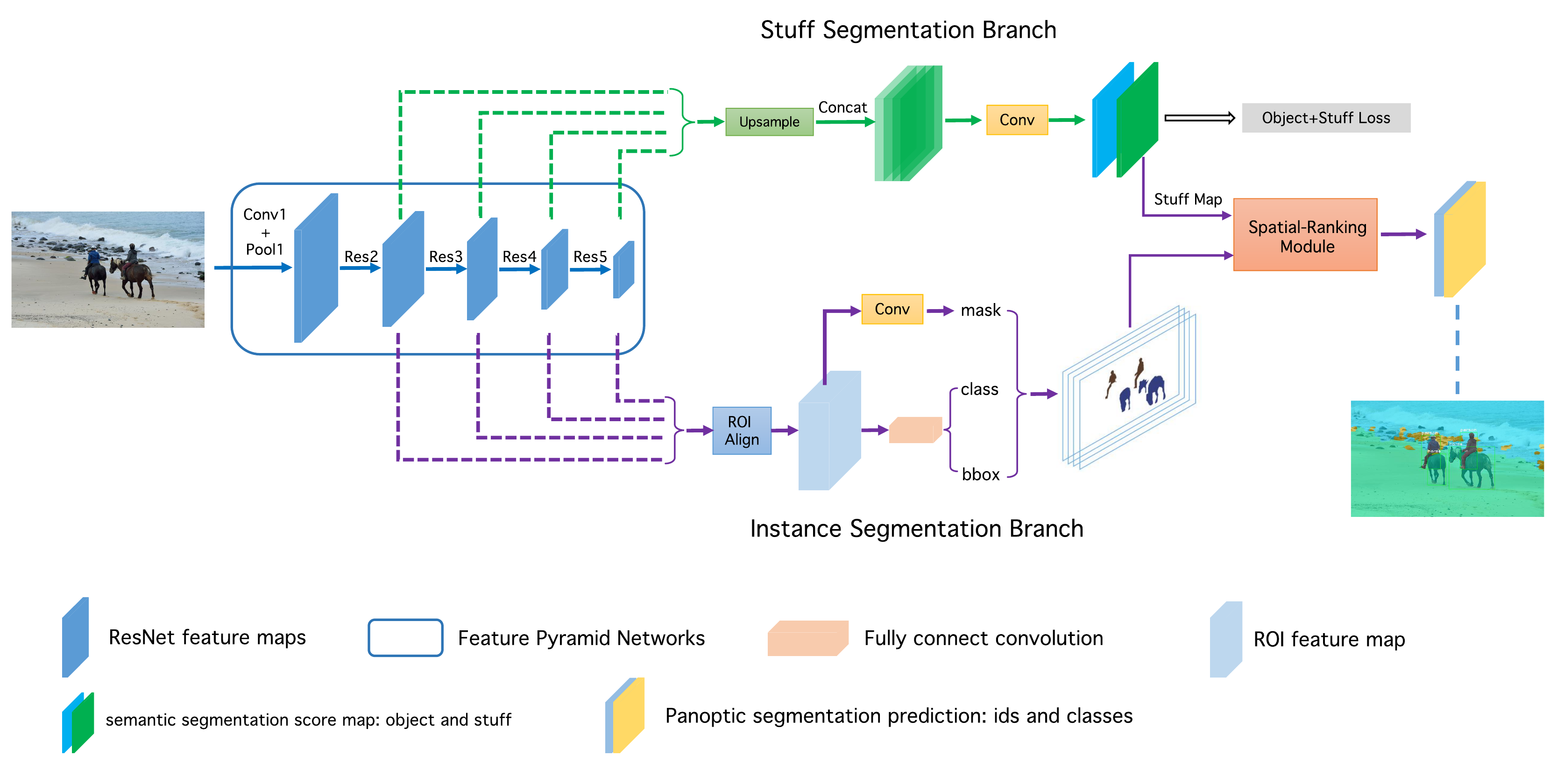}
		\end{minipage}
	}
	\caption{The illustration of overall framework. Given an input image, we use the FPN network to provide feature maps for stuff branch and instance branch. The two branches generate intermediate results, while later are passed to our spatial ranking module. The spatial ranking module learns a ranking score for each instance as the final merging evidence. }
	\label{fig:network_stucture}
\end{figure*}

\section{Proposed End-to-end Framework}
The overview of our algorithm is illustrated in Figure \ref{fig:network_stucture}. There are three major components in our algorithm: 1) The stuff branch predicts the stuff segmentation for the whole input. 2) The instance branch provides the instance segmentation predictions. 3) The spatial ranking module generates a ranking score for the each instance.

\subsection{End-to-end Network Architecture}
We employ FPN~\cite{lin2017feature} as the backbone architecture for the end-to-end network. For instance segmentation, we adopt the original Mask R-CNN~\cite{he2017mask} as our network framework. We apply top-down pathway and lateral connections to get feature maps. And then, a $3\times3$ convolution layer is appended to get RPN feature maps. After that, we apply the ROIAlign~\cite{he2017mask} layer to extract object proposal features and get three predictions: proposal classification score, proposal bounding box coordinates, and proposal instance mask.

\begin{figure}
	\subfigure{
		\begin{minipage}{0.95\textwidth}
		\label{main figure}
		\includegraphics[width=0.5\linewidth]{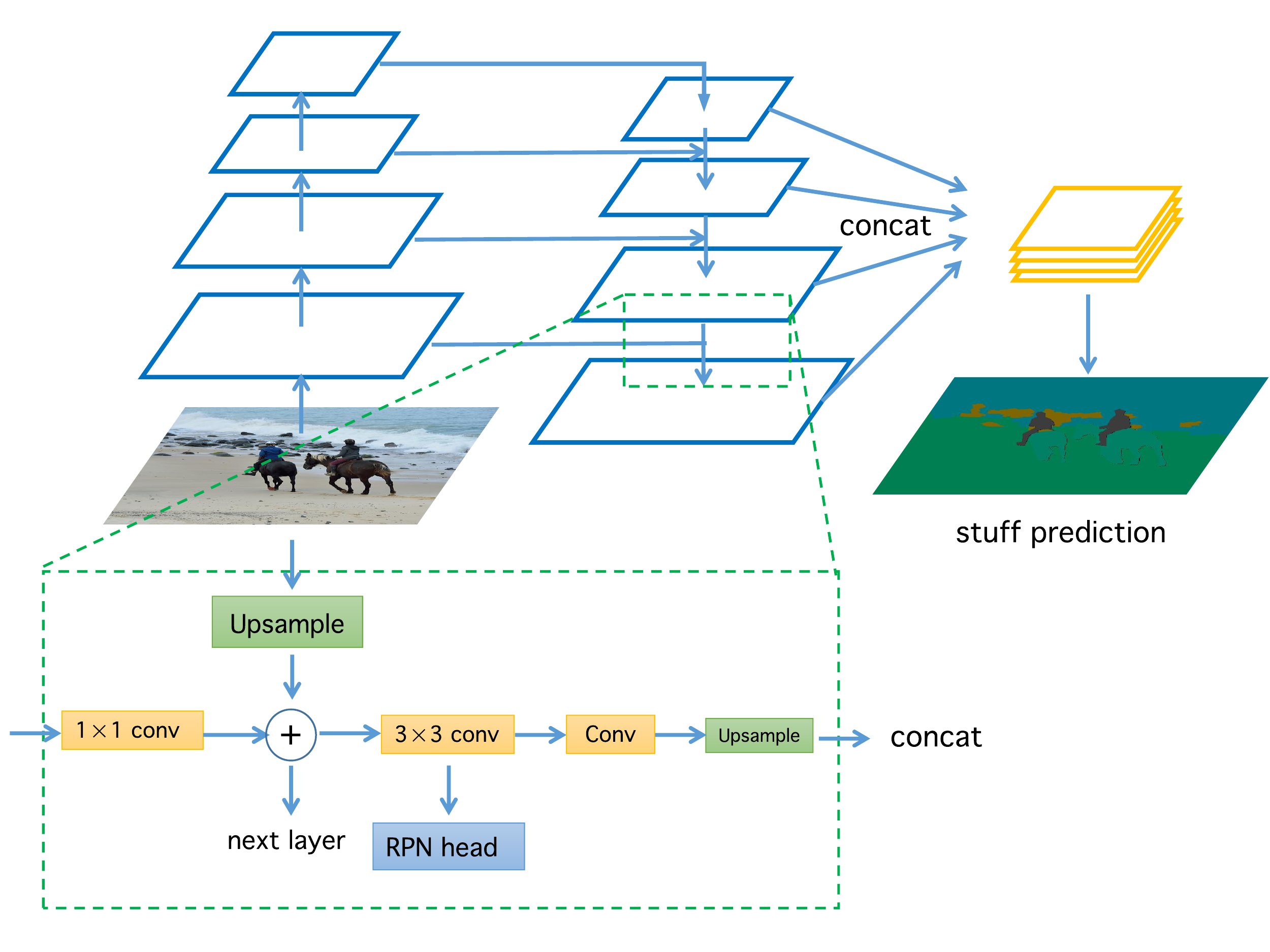}
		\end{minipage}
	}
	\caption{A building block illustrating the stuff segmentation subnetwork. Here we share both the backbone and skip-connection feature maps in stuff branch and instance branch. Besides, we predict both object and stuff category for the stuff branch. }
	\label{fig:stuff_subnetwork_stucture}
\end{figure}

For stuff segmentation, two $3\times3$ convolution layers are stacked on RPN feature maps. For the sake of multi-scale feature extraction, we then concatenate these layers with succeeding one $3\times3$ convolution layer and $1\times1$ convolution layer. Figure~\ref{fig:stuff_subnetwork_stucture} presents the details of stuff branch. During training, we supervise the stuff segmentation and thing segmentation simultaneously, as the auxiliary objection information could provide object context for stuff prediction. In inference, we only extract the stuff predictions and normalized them to probability.

To break out the information flow barrier during training and to make the whole pipeline more efficient, we share the features from the backbone network of two branches. The issue raised here could be divided into two parts: 1) the sharing granularity on feature maps and 2) the balance between instance loss and stuff loss. In practice, we find that as more feature maps are shared, better performance we can obtain. Thus, we share the feature maps until skip-connection layers, that is the $3\times 3$ convolution layer before RPN head shown in Figure~\ref{fig:stuff_subnetwork_stucture}.
\begin{equation}
\begin{split}
L_{\text{total}} =
&  \underbrace{L_{\text{rpn\_cls}} + L_{\text{rpn\_bbox}} + L_{\text{cls}} + L_{\text{bbox}} + L_{\text{mask}}}_{\text{instance branch}} \\
& + \underbrace{\lambda \cdot L_{\text{seg}\_{(\text{stuff+object)}}}}_{\text{stuff branch}} \; + \; L_{\text{srm}}
\end{split}
\label{equation_loss_balance}
\end{equation}

\begin{figure*}
	\subfigure{
		\begin{minipage}{1.0\textwidth}
		\label{main figure}
		\includegraphics[width=1\linewidth]{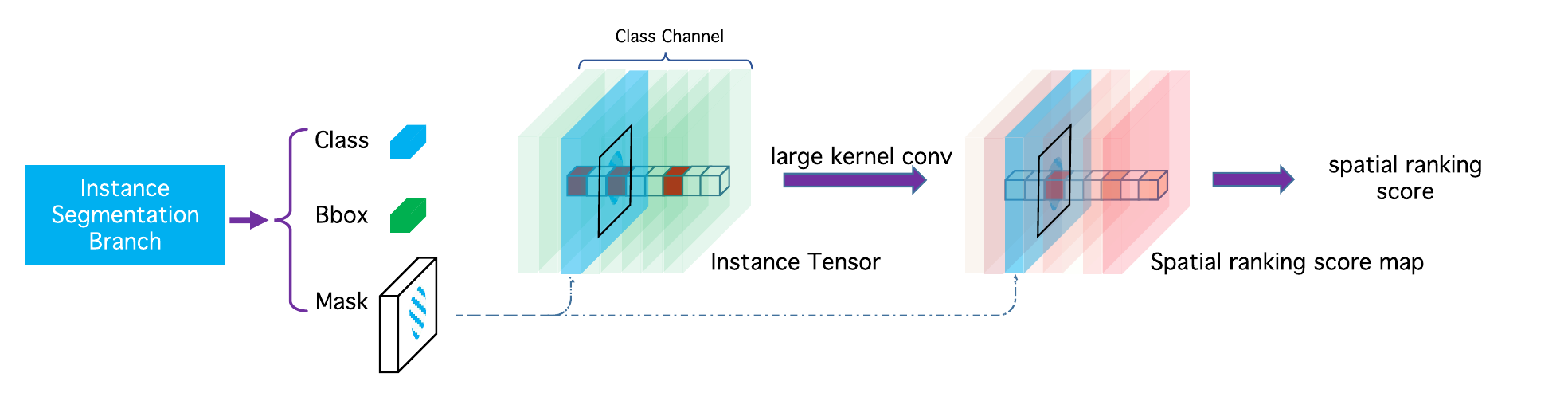}
		\end{minipage}
	}
	\caption{An illustration of spatial ranking score map prediction. The pixel vector in instance feature map represents instance prediction result in this pixel. The red color means that the corresponding category object includes this pixel and the multiple red channels indicate the occlusion problem between instances. We use the panoptic segmentation category label to supervise spatial ranking score map.}
	\label{fig:spatial_ranking_module}
\end{figure*}

As for the balance of two supervisions, we first present the multiple losses in Equation~\ref{equation_loss_balance}. The instance branch contains 5 losses: $L_{rpn\_cls}$ is the RPN objectness loss, $L_{rpn\_bbox}$ is the RPN bounding-box loss, $L_{cls}$ is the classification loss, $L_{bbox}$ is the object bounding-box regression loss, and $L_{mask}$ is the average binary cross-entropy loss for mask prediction. As the stuff branch, there is only one semantic segmentation loss named $L_{seg\_(stuff+object)}$. The hyperparameter $\lambda$ is employed for loss balance and will be discussed later. $L_{\text{srm}}$ represents the loss function of the spatial ranking module, which is described in the next section.

\subsection{Spatial Ranking Module}
The modern instance segmentation framework is often based on object detection network with an additional mask prediction branch, such as the Mask RCNN~\cite{he2017mask} which is usually based on FPN~\cite{lin2017feature}. Generally speaking, the current object detection framework does not consider the overlapping problems between different classes, since the popular metrics are not affected by this issue, e.g., the AP and AR. However, in the task of panoptic segmentation, since the number of pixels in one image is fixed, the overlapping problem, or specifically the multiple assignments for one pixel, must be resolved.

Commonly, the detection score was used to sort the instances in descending order, and then assign them to the stuff canvas by the rule of larger score objects on top of lower ones. However, this heuristic algorithm could easily fail in practice. For example, let's consider a person wearing a tie, shown in Figure~\ref{fig:visual_srm_example}. As the person class is more frequent than the tie in the COCO dataset, its detection score is tend to be higher than the tie bounding box. Thus through the above simple rule, the tie instance is covered by the person instance, and leading to the performance drops.

Could we alleviate this phenomenon through the panoptic annotation? That is if we force the network learns the person annotation with a hole in the place of the tie, could we avoid the above situation? As shown in Table~\ref{tab_shr}, we conduct the experiment with the above mentioned annotations, but only find the decayed performance. Therefore, this approach is not applicable currently.

To counter this problem, we resort to a semantic-like approach and propose a simple but very effective algorithm for dealing with occlusion problems, called \emph{spatial ranking module}. As shown in the Figure~\ref{fig:spatial_ranking_module}, we first map the results of the instance segmentation to the tensor of input size.  The dimension of the feature map is the number of object categories, and the instances of different categories are mapped to the corresponding channels. The instance tensor is initialized to zero, and the mapping value is set to one. We then append the large kernel convolution~\cite{peng2017large} after the tensor to obtain the ranking score map. In the end , we use the pixel-wise cross entropy loss to optimize the ranking score map, as the Equation~\ref{equation_loss_srm} shows. $S_{map}$ represents the output ranking score map and $S_{label}$ represents the corresponding non-overlap semantic label.
\begin{equation}
L_{\text{srm}} = CE(S_{map}, S_{label})
\label{equation_loss_srm}
\end{equation}
After getting the ranking score map, we calculate the ranking score of each instance object as Equation~\ref{eq:ranking_score}. Here, $S_{i,j,cls}$ represents the ranking score value in $(i, j)$ of class $cls$, note that $S_{i,j,cls}$  has been normalized to a probability distribution. $m_{i,j}$ is the mask indicator, representing if pixel $(i, j)$ belongs to the instance. The ranking score of the whole instance $P_{objs}$ is computed by the average of pixel ranking scores in a mask.
\begin{equation}
    P_{objs} = \frac{\sum_{(i,j)\in objs} S_{i,j,cls} \cdot m_{i,j}}{\sum_{(i,j)\in objs} m_{i,j}}
    \label{eq:ranking_score}
\end{equation}
\begin{equation}
m_{i,j}=
\begin{cases}
0& \text{(i,j) $\in$ instance}\\
1& \text{(i,j) $\notin$ instance}
\end{cases}
\end{equation}
Let's reconsider the example mentioned above through our proposed spatial ranking module. When we forward the person mask and tie mask into this module, we obtain the spatial ranking scores using Equation~\ref{eq:ranking_score} for these two objects. Within the ranking score, the sorting rule in the previous method could be more reliable, and the performance is improved, as the experiments present in the next section.
\section{Experiments}

\subsection{Dataset and Evaluation Metrics}

\textbf{Dataset:}
We conduct all experiments on COCO panoptic segmentation dataset~\cite{kirillov2018panoptic}. This dataset contains 118K images for training, 5k images for validation, with annotations on 80 categories for the thing and 53 classes for stuff. We only employ the training images for model training and test on the validation set. Finally, we submit test-dev result to the COCO 2018 panoptic segmentation leaderboard.

\textbf{Evaluation metrics:}
We use the standard evaluation metric defined in~\cite{kirillov2018panoptic}, called Panoptic Quality (PQ). It contains two factors: 1) the Segmentation Quality (SQ) measures the quality of all categories and 2) the Detection Quality (DQ) measures only the instance classes. The mathematical formations of PQ, SQ and DQ are presented in Equation~\ref{eq:pq_sq_dq_math}, where p and g are predictions and ground truth, and TP, FP, FN represent true positives, false positives and false negatives. It is easy to find that SQ is the common mean IOU metric normalized for matching instances, and DQ could be regarded as a form of detection accuracy. The matching threshold is set to 0.5, that is if the pixel IOU of prediction and ground truth is larger than 0.5, the prediction is regarded matched, otherwise unmatched. For stuff classes, each stuff class in an image is regarded as one instance, no matter the shape of it.

\begin{equation}
    PQ = \underbrace{\frac{\sum_{(p,g)\in TP}IOU(p,g)}{\left| TP \right|}}_{\text{Segmentation Quality (SQ)}} \times \underbrace{\frac{\left| TP \right|}{\left| TP \right| + \frac{1}{2}\left| FP \right| + \frac{1}{2}\left| FN \right|}}_{\text{Detection Quality (DQ)}}
    \label{eq:pq_sq_dq_math}
\end{equation}

\subsection{Implementation Details}
We choose the ResNet-50~\cite{he2016deep} pretrained on ImageNet for ablation studies. We use the SGD as the optimization algorithm with momentum $0.9$ and weight decay $0.0001$. The multi-stage learning rate policy with warm up strategy~\cite{peng2018megdet} is adopted. That is, in the first $2,000$ iterations, we use the linear gradual warmup policy by increasing the learning rate from 0.002 to $0.02$. After $60,000$ iterations, we decrease the learning rate to $0.002$ for the next $20,000$ iterations and further set it to $0.0002$ for the rest $20,000$ iterations. The batch size of input is set to $16$, which means each GPU consumes two images in one iteration. For other details, we employ the experience from Mask-RCNN~\cite{he2017mask}.

Besides the training for the two branches of our network, a little bit more attention should be paid to the spatial ranking module. 
During the training process, the supervision label is the
corresponding non-overlap semantic label and training it
as a semantic segmentation network. We set the non-conflicting
pixels ignored to force the network focus on the
conflicting regions.

During the inference, we set the max number of boxes for each image to 100 and the min area of a connected stuff area to 4,900. As for the spatial ranking module, since we do not have the ground truth now, the outputs of instance branch will be passed through this module to resolve the overlapping issue.

\subsection{Ablation Study on Network Structure}
In this subsection, we focus on the properties of our end-to-end network design. There are three points should be discussed as follows: the loss balance parameter, the object context for stuff branch and the sharing mode of two branches. To avoid the Cartesian product of experiments, we only modify the specific parameters and control the other optimally.

\begin{figure}[htbp]
	\subfigure{
		\begin{minipage}{1.0\textwidth}
		\label{main figure}
		\includegraphics[width=0.5\linewidth]{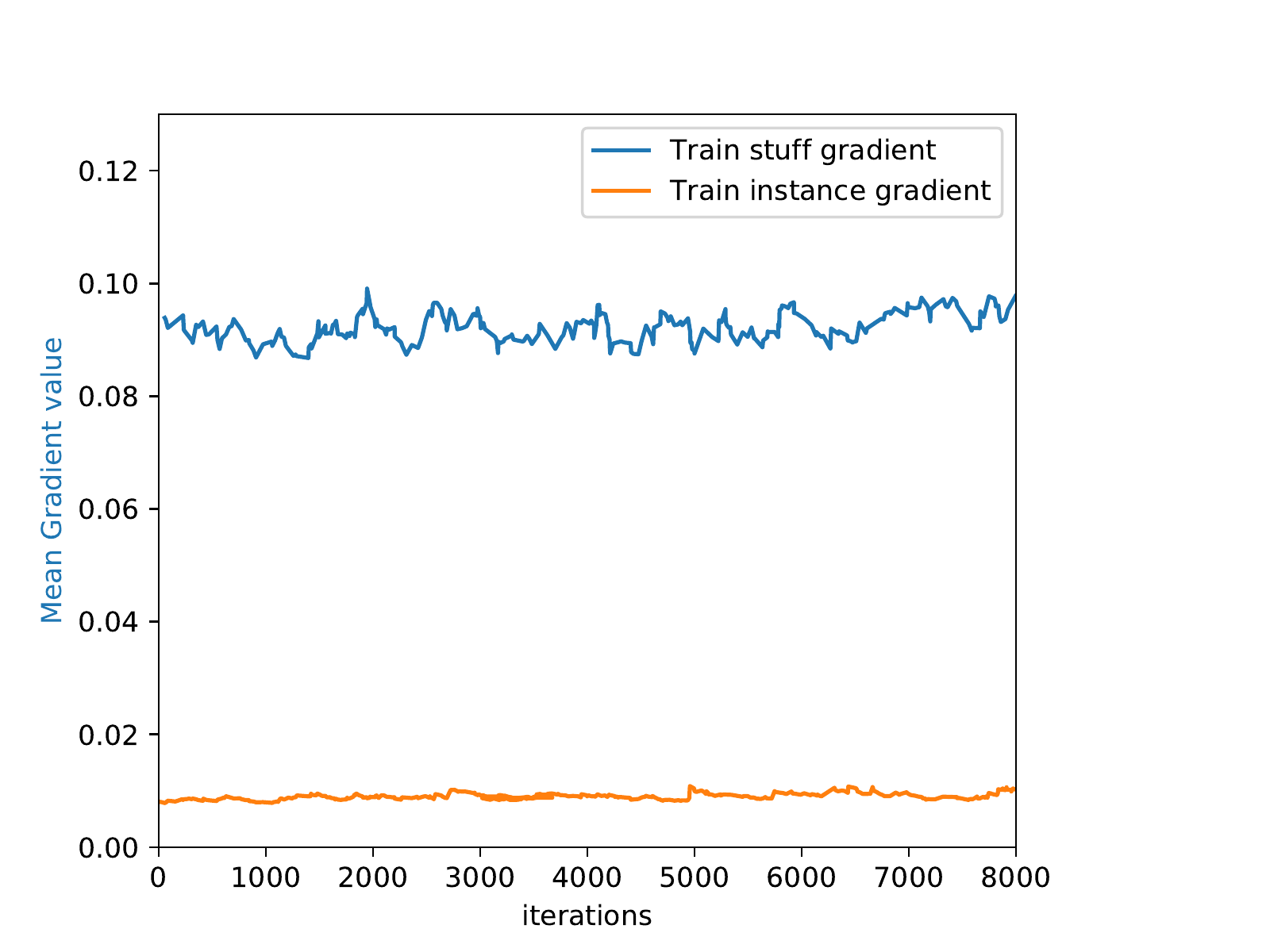}
		\end{minipage}
	}
	\caption{The plot of the specified layer mean gradient value in two branches. We chose one epoch iterations in the training process. The learning rate of the two branches is the same. The horizontal axis is the number of iterations, and the vertical axis is the mean gradient value of the backbone last layer.}
	\label{fig:gradient_balance}
\end{figure}

\begin{table}[htbp]
\begin{center}
\scalebox{1}{
\begin{tabular}{c|ccc}
\hline
$\lambda$ &\text{PQ} &$\text{PQ}^\text{Th}$ &$\text{PQ}^\text{St}$  \\
\hline
\hline
0.2 &36.9 &45.0  &24.6 \\
0.25 &\textbf{37.2} &45.4  &24.9 \\
0.33 &36.9 &44.4  &25.4 \\
0.50 &36.5 &43.5 &25.9  \\
0.75 &35.3 &41.9  &25.4 \\
1.0 &- &- &- \\

\hline
\end{tabular}
}
\end{center}
\caption{Loss balance between instance segmentation and stuff segmentation}
\label{feature_loss_balance}
\end{table}

\begin{figure*}[t]
\centering
\subfigure{
\begin{minipage}[t]{0.30\linewidth}
\includegraphics[height=0.5\linewidth,width=1.02\linewidth]{basemodel_sharing_000000209747_image.png}\vspace{1pt}
\includegraphics[height=0.6\linewidth,width=1.02\linewidth]{basemodel_sharing_000000214753_image.png}\vspace{1pt}
\includegraphics[height=0.6\linewidth,width=1.02\linewidth]{basemodel_sharing_000000211825_image.png}\vspace{1pt}
\includegraphics[height=0.8\linewidth,width=1.02\linewidth]{basemodel_sharing_000000213171_image.png}\vspace{1pt}
\centerline{Input Image}

\end{minipage}}
\subfigure{
\begin{minipage}[t]{0.30\linewidth}
\includegraphics[height=0.5\linewidth,width=1.02\linewidth]{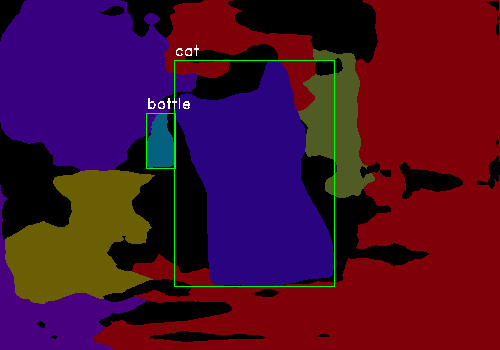}\vspace{1pt}
\includegraphics[height=0.6\linewidth,width=1.02\linewidth]{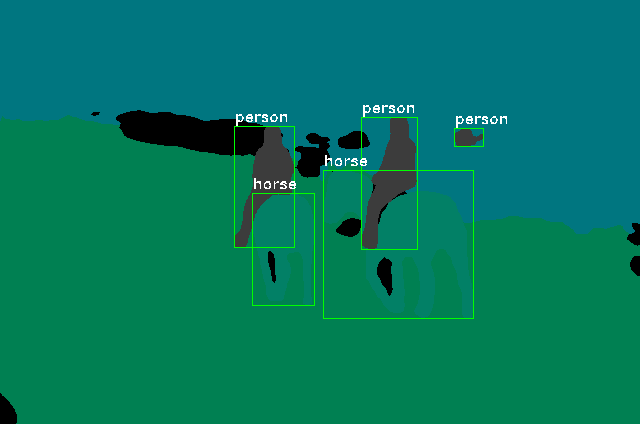}\vspace{1pt}
\includegraphics[height=0.6\linewidth,width=1.02\linewidth]{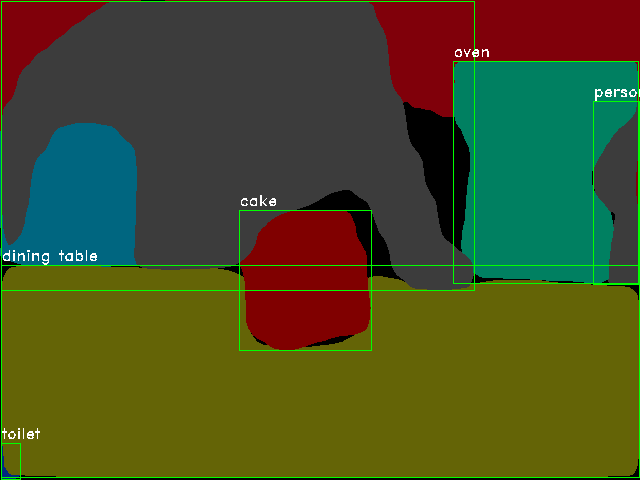}\vspace{1pt}
\includegraphics[height=0.8\linewidth,width=1.02\linewidth]{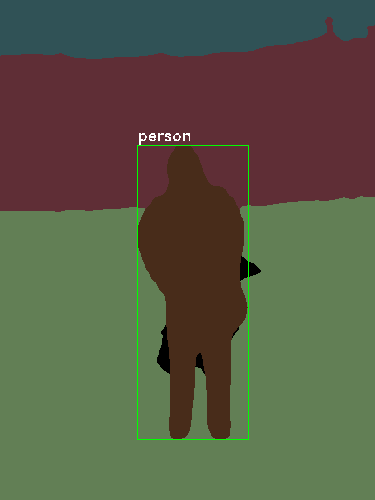}\vspace{1pt}
\centerline{No-sharing Result}

\end{minipage}}
\subfigure{
\begin{minipage}[t]{0.30\linewidth}
\includegraphics[height=0.5\linewidth,width=1.02\linewidth]{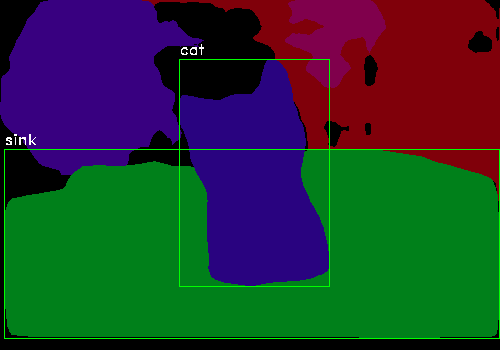}\vspace{1pt}
\includegraphics[height=0.6\linewidth,width=1.02\linewidth]{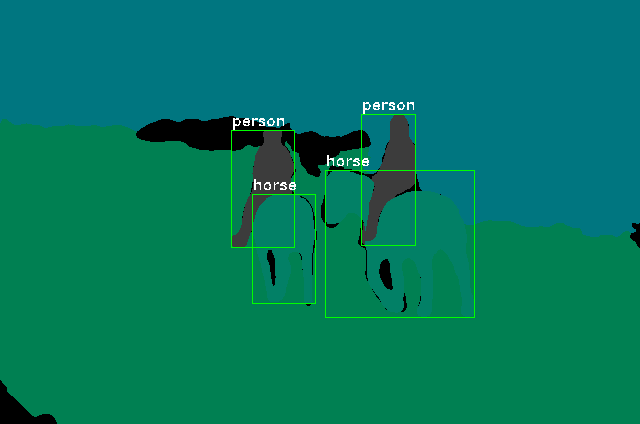}\vspace{1pt}
\includegraphics[height=0.6\linewidth,width=1.02\linewidth]{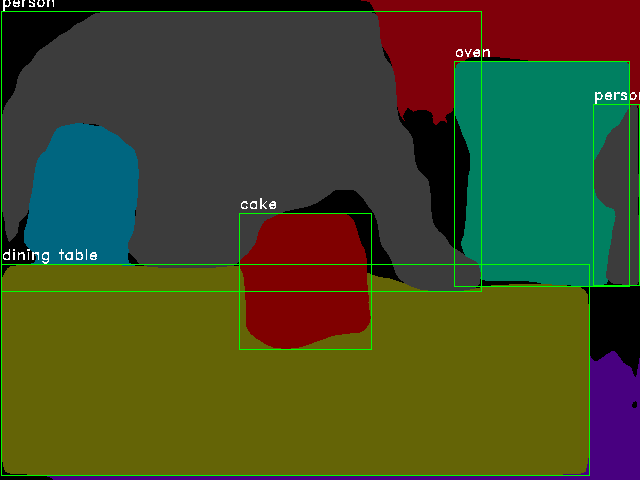}\vspace{1pt}
\includegraphics[height=0.8\linewidth,width=1.02\linewidth]{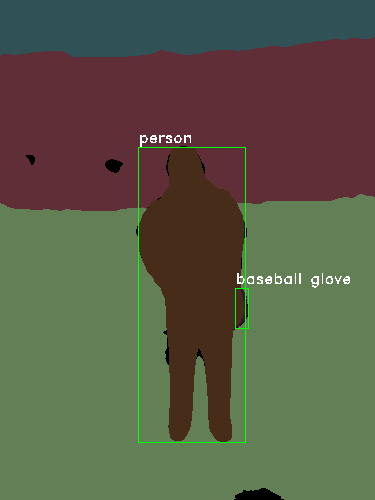}\vspace{1pt}
\centerline{Sharing Result}

\end{minipage}}

\vspace{3pt}
\caption{Feature Sharing Mode Visualization. The first column is the original image, and the second column is sharing backbone feature result, and the last column is the no-sharing result.}
\label{fig:figure_sharing_feature}
\end{figure*}

\textbf{The loss balance} issue comes from the reality that the gradients from stuff branch and instance branch are not close. We make a statistic on the mean gradients of two branches with respect to the last feature map of the backbone, where the hyperparameter $\lambda$ is set to 1 for fairness.

As shown in the Figure~\ref{fig:gradient_balance}, it is brief and clear that the gradient from stuff branch dominates the penalty signals. Therefore, we obtain a hyperparameter $\lambda$ in Equation~\ref{equation_loss_balance} to balance the gradients.  We conduct the experiments with $\lambda \in [0.2, 0.25, 0.33, 0.5, 0.75, 1.0]$. The interval is not uniform for the sake of searching efficiency. As the Table~\ref{feature_loss_balance} summarizes, $\lambda=0.25$ is the optimal choice. Note that if we set $\lambda=1$, which means the instance and stuff branches trained like separate models, the network could not converge through our default learning rate policy.

\textbf{Object context} is a natural choice in stuff segmentation. Although we only want the stuff predictions from this branch, the lack of object supervision will introduce holes on ground truth, resulting in discontinuous context around objects. Therefore, we conduct a pair of comparative experiments where all 133 categories is supervised , and the other is trained on 53 stuff classes. The results in Table~ \ref{table_add_object_loss} shows the 0.5 improvement on overall PQ with object context.
\begin{table}
\begin{center}
\scalebox{1}{
\begin{tabular}{c|c|ccc}
\hline
Stuff-SC &Object-SC &\text{PQ} &$\text{PQ}^\text{Th}$ &$\text{PQ}^\text{St}$  \\
\hline
\hline
\checkmark &- &$36.7$ &$43.8$ &$25.9$ \\
\checkmark &\checkmark &$\textbf{37.2}$ &$45.4$ &$24.9$ \\

\hline
\end{tabular}
}
\end{center}
\caption{Ablation study results on stuff segmentation network design. Stuff-SC represents stuff supervision classes. It refers to predict stuff classes. While both Stuff-SC and Object-SC mean predicting all classes. }
\label{table_add_object_loss}
\end{table}

\begin{table*}
\begin{center}
\scalebox{1.0}{
\begin{tabular}{c|c|ccc|ccc|ccc}
\hline
Methods &backbone &\text{PQ} &$\text{PQ}^\text{Th}$ &$\text{PQ}^\text{St}$ &\text{SQ} &$\text{SQ}^\text{Th}$ &$\text{SQ}^\text{St}$ &\text{DQ} &$\text{DQ}^\text{Th}$ &$\text{DQ}^\text{St}$ \\
\hline
\hline
baseline &ResNet-50 &$37.2$ &$45.4$ &$24.9$ &$77.1$ &$81.5$ &$70.6$ &$45.7$ &$54.4$ &$32.5$ \\
w/pano-instance GT &ResNet-50 &36.1 &43.5 &24.9 &76.1 &80.0 &70.3 &44.5 &52.4 &32.7 \\
w/spatial ranking module &ResNet-50 &$\textbf{39.0}$ &$48.3$ &$24.9$ &$77.1$ &$81.4$ &$70.6$ &$47.8$ &$58.0$ &$32.5$ \\

\hline
\hline
baseline &ResNet-101 &$38.8$ &$46.9$ &$26.6$ &$78.2$ &$82.0$ &$72.5$ &$47.4$ &$55.9$ &$34.5$ \\
w/spatial ranking module &ResNet-101 &$\textbf{40.7}$ &$50.0$ &$26.6$ &$78.2$ &$82.0$ &$72.5$ &$49.6$ &$59.7$ &$34.5$ \\

\hline
\end{tabular}
}
\end{center}
\caption{Results on MS-COCO panoptic segmentation validation dataset which use our spatial ranking module method. W/pano-instance GT represents using panoptic segmentation ground truth to generate instance segmentation ground truth. It is trained in two separate networks. All results in this table are based on backbone ResNet-50.}
\label{tab_shr}
\end{table*}

\textbf{Sharing features} is a key point of our network design. The benefits of sharing have two parts: 1) two branches could absorb useful information from other supervision signals, and 2) the computation resources could be saved if the shared network is computed only once. To investigate the granularity on sharing features, we conduct two experiments, where the \emph{shallow share model} only shares the backbone features and the \emph{deep share model} further shares the feature maps before RPN head, as Figure~\ref{fig:stuff_subnetwork_stucture} presents. Table~\ref{share_feature_or_not} shows the comparisons between different settings, and \emph{deep share model} outperform the separate training baseline by 0.7\% on PQ. Figure~\ref{fig:figure_sharing_feature} presents the visualization of sharing features.



\begin{table}
\begin{center}
\scalebox{1}{
\begin{tabular}{c|c|ccc}
\hline
\text{Methods} &\text{backbone} &\text{PQ} &$\text{PQ}^\text{Th}$ &$\text{PQ}^\text{St}$ \\
\hline
\hline
no &ResNet-50 &$36.5$ &$44.4$ &$24.6$ \\
res1-res5 &ResNet-50 &$37.0$ &$44.8$ &$25.2$  \\
+ skip-conection &ResNet-50 &$\textbf{37.2}$ &$45.4$ &$24.9$ \\

\hline
\hline
no &ResNet-101 &$38.2$ &$46.3$ &$26.0$ \\

+ skip-conection &ResNet-101 &$\textbf{38.8}$ &$46.9$ &$26.6$ \\

\hline
\end{tabular}
}

\end{center}
\caption{Results on whether share stuff segmentation and instance segmentation features. On ResNet-50 backbone, sharing features method gets a gain of 0.7 in PQ, and ResNet-101 gets a gain of 0.7. Ablation study results on different sharing feature way. The res1-res5 means just share the backbone ResNet features. The +skip-connection means share both the backbone features and FPN skip-connection branch.}
\label{share_feature_or_not}

\end{table}

\begin{table}
\begin{center}
\scalebox{1}{
\begin{tabular}{c|ccc}
\hline
\text{Conv Settings} &\text{PQ} &$\text{PQ}^\text{Th}$ &$\text{PQ}^\text{St}$ \\
\hline
\hline
$1 \times 1$ &$38.4$ &$47.4$ &$24.9$ \\
$3 \times 3$ &$38.7$ &$47.8$ &$24.9$ \\
$1 \times 7$ + $7 \times 1$&\textbf{39.0} &$48.3$ &$24.9$ \\
\hline
\end{tabular}
}

\end{center}
\caption{Results on the convolution settings of spatial ranking module. $1 \times 1$ represents the convolution kernel size is 1. Results shows that the large receptive field can help the spatial ranking module get more context features and better results.}
\label{srm_conv_setting}

\end{table}

\subsection{Ablation Study on Spatial Ranking Module}

\textbf{Supervised by no-overlapping annotations} are a straightforward idea to resolve the object ranking issue. Here, we process the panoptic ground truth and extract the non-overlapping annotations for instance segmentation. The 3rd row of Table~\ref{tab_shr} gives the results of this idea. Unfortunately, merely replacing the instance ground truth do not help improve the performance, and may reversely reduce the accuracy and recall for objects. This phenomenon may come from the fact that most of the objects in COCO do not meet the overlapping issue, and forcing the network to learn non-overlapping hurts the overall performance.

\begin{table}
\begin{center}
\scalebox{1}{
\begin{tabular}{c|ccc}
\hline
\text{Methods} &\text{PQ} &$\text{PQ}^\text{Th}$ &$\text{PQ}^\text{St}$ \\
\hline
\hline
Artemis &$16.9$ &$16.8$ &$17.0$ \\
JSIS-Net~\cite{de2018panoptic} &$27.2$ &$29.6$ &$23.4$ \\
MMAP-seg &$32.1$ &$38.9$ &$22.0$ \\
LeChen &$37.0$ &$44.8$ &$25.2$  \\
\hline
Ours(OANet) &\textbf{41.3} &$50.4$ &$27.7$ \\
\hline
\end{tabular}
}

\end{center}
\caption{Results on the COCO 2018 panoptic segmentation challenge test-dev. Results verifies the effectiveness of our feature sharing mode and the spatial ranking module. We use the ResNet-101 as our basemodel.}
\label{coco_test_dev_result}

\end{table}

\begin{figure}[htbp]
	\subfigure{
		\begin{minipage}{1.0\textwidth}
		\label{main figure}
		\includegraphics[width=0.5\linewidth]{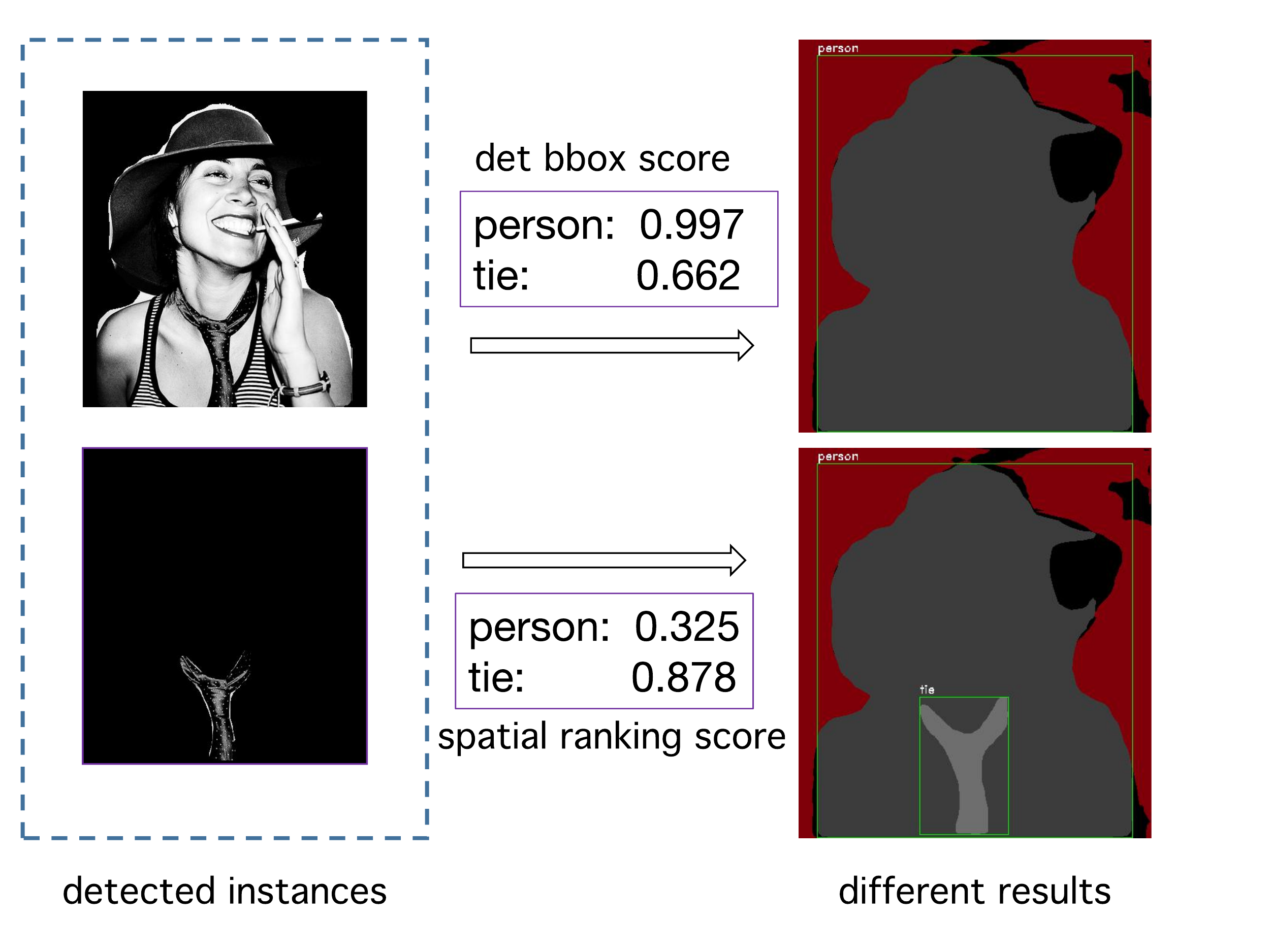}
		\end{minipage}
	}
	\caption{The visualization result of the spatial ranking module. The left two segments denote the detected instances through our model, the det bbox score represents the object detection score predicted in detection. The spatial ranking score represents the values from our approach. }
	\label{fig:visual_srm_example}
\end{figure}

\begin{figure*}[t]
\centering
\subfigure{
\begin{minipage}[t]{0.30\linewidth}
\includegraphics[height=0.6\linewidth,width=1.02\linewidth]{srm_000000003845_image}\vspace{1pt}
\includegraphics[height=0.6\linewidth,width=1.02\linewidth]{srm_000000006040_image}\vspace{1pt}
\includegraphics[height=0.6\linewidth,width=1.02\linewidth]{srm_000000010764_image}\vspace{1pt}
\includegraphics[height=0.6\linewidth,width=1.02\linewidth]{srm_000000011051_image}\vspace{1pt}
\centerline{Input Image}

\end{minipage}}
\subfigure{
\begin{minipage}[t]{0.30\linewidth}
\includegraphics[height=0.6\linewidth,width=1.02\linewidth]{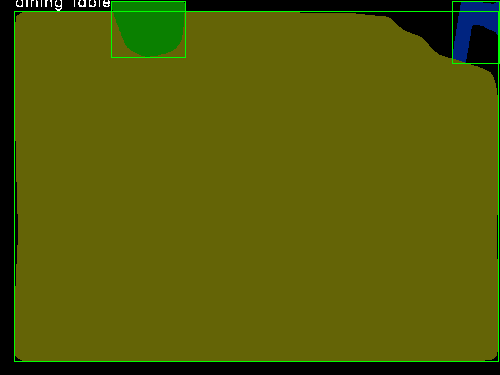}\vspace{1pt}
\includegraphics[height=0.6\linewidth,width=1.02\linewidth]{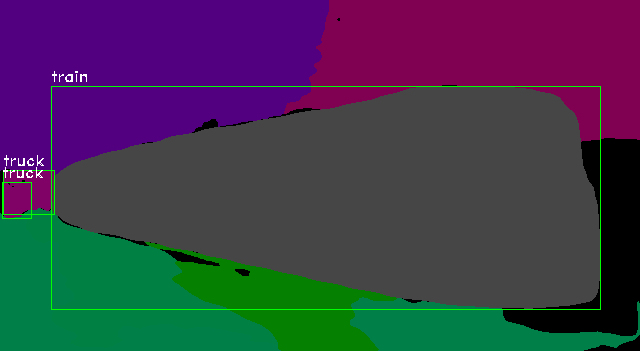}\vspace{1pt}
\includegraphics[height=0.6\linewidth,width=1.02\linewidth]{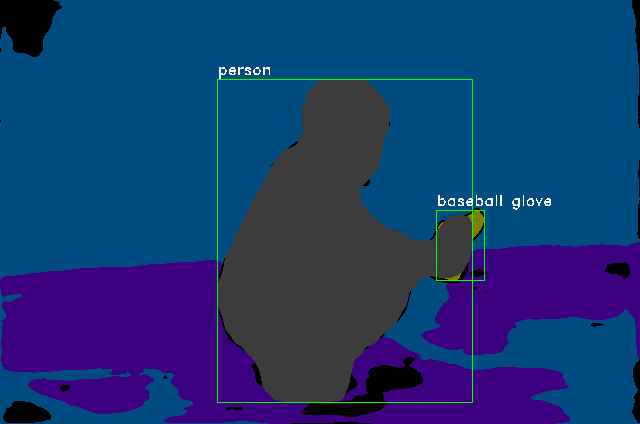}\vspace{1pt}
\includegraphics[height=0.6\linewidth,width=1.02\linewidth]{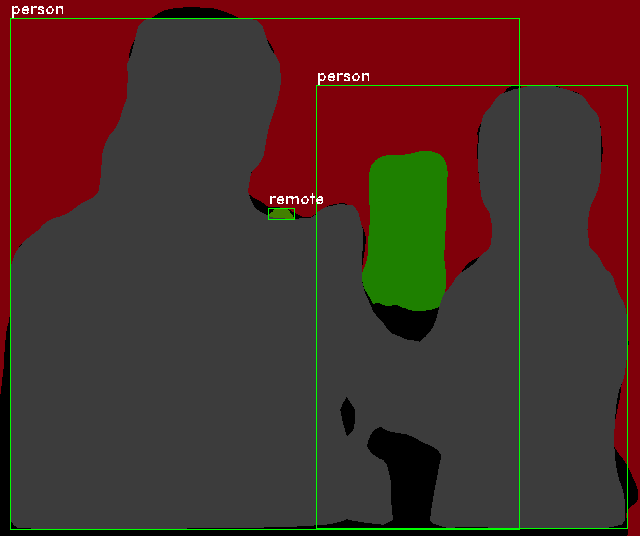}\vspace{1pt}
\centerline{Heuristic Method}
\end{minipage}}
\subfigure{
\begin{minipage}[t]{0.30\linewidth}
\includegraphics[height=0.6\linewidth,width=1.02\linewidth]{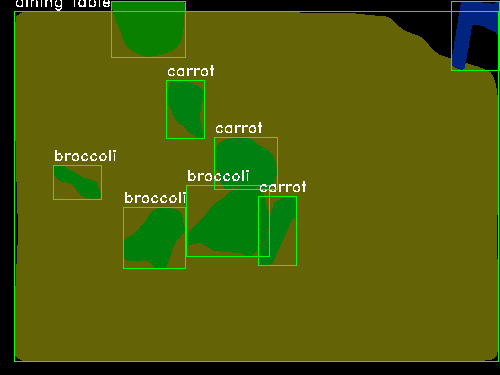}\vspace{1pt}
\includegraphics[height=0.6\linewidth,width=1.02\linewidth]{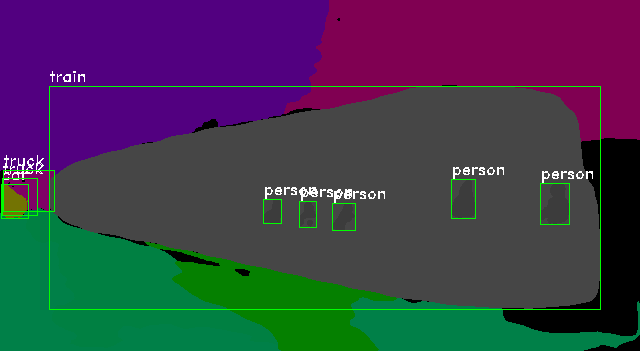}\vspace{1pt}
\includegraphics[height=0.6\linewidth,width=1.02\linewidth]{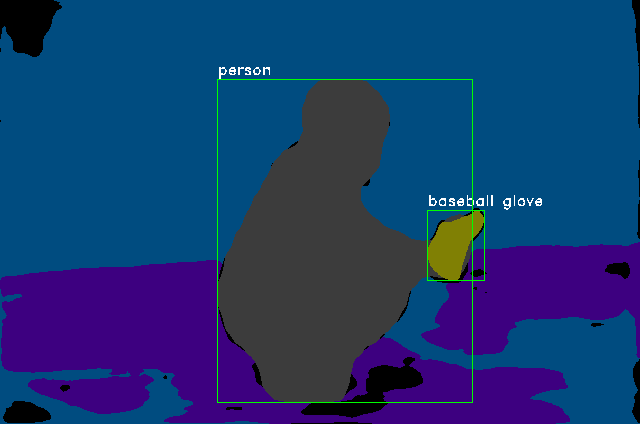}\vspace{1pt}
\includegraphics[height=0.6\linewidth,width=1.02\linewidth]{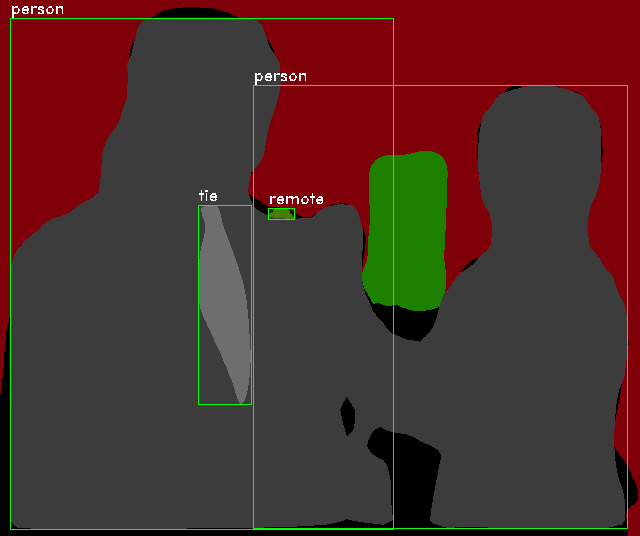}\vspace{1pt}
\centerline{Our Method}

\end{minipage}}

\vspace{3pt}
\caption{The visualization results using our spatial ranking module. The first column is the input image, and second column is the panoptic segmentation result with heuristic method, and last column is the result using our approach.}
\label{figure_shr}
\end{figure*}

\parskip=0pt

\textbf{Spatial ranking module} proposed in this paper is aimed to solve the overlapping issue in panoptic segmentation. Table~\ref{srm_conv_setting} shows that large receptive field can help the spatial ranking module get more context features and better results.
As we can see in 3rd row or 2nd row for the Resnet101 of Table~\ref{tab_shr}, compared with the above end-to-end baseline, our spatial ranking module improve the PQ by 1.8\%. Specifically, the $\text{PQ}^{\text{Th}}$ is increased by 2.9\%, while the metrics for stuff remains the same. These facts prove the purpose of our spatial ranking module is justified. We test our OANet on COCO test-dev, as shown in Table~\ref{coco_test_dev_result}. Compared with the results of other methods, our method achieves the state-of-the-art result. For detailed results, please refer to the leardboard~\footnote[3]{http://cocodataset.org/\#panoptic-leaderboard}.

\parskip=0pt
Figure~\ref{fig:visual_srm_example} explaines the principle of our spatial ranking module. For the example input image, the network predicts a person plus a tie, and their bounding box scores are 0.997 and 0.662 respectively. If we use the scores to decide the results, the tie will definitely be covered by the person. However, in our method, we can get a spatial ranking score for each instance, 0.325 and 0.878 respectively. With the help of the new scores, we can get the right predictions. Figure~\ref{figure_shr} summarizes more examples.

\section{Conclusion}
In this paper, we propose a novel end-to-end occlusion aware algorithm, which incorporates the common semantic segmentation and instance segmentation into a single model. In order to better employ the different supervisions and reduce the consumption of computation resources, we investigate the feature sharing between different branches and find that we should share as many features as possible. Besides, we have also observed the particular ranking problem raised in the panoptic segmentation, and design the simple but effective spatial ranking module to deal with this issue. The experiment results show that our approach outperforms the previous state-of-the-art models.
{\small
\bibliographystyle{ieee}
\bibliography{egbib}
}

\end{document}